\documentclass{article} 
\usepackage{iclr2026_conference,times}

\usepackage{amsmath,amsfonts,bm}

\def\eqref#1{equation~\ref{#1}}

\def\1{\bm{1}}

\DeclareMathAlphabet{\mathsfit}{\encodingdefault}{\sfdefault}{m}{sl}
\SetMathAlphabet{\mathsfit}{bold}{\encodingdefault}{\sfdefault}{bx}{n}

\usepackage{hyperref}
\usepackage{url}
\usepackage[nice]{nicefrac}
\usepackage{amsmath}
\usepackage{amssymb}
\usepackage{mathtools}
\usepackage{algorithm}
\usepackage{algpseudocode}
\usepackage{graphicx}
\usepackage{wrapfig}
\usepackage{multirow}
\usepackage{booktabs}
\usepackage[inline]{enumitem}
\usepackage{setspace}
\usepackage{subcaption}
\usepackage[noEnd=false, indLines=true]{algpseudocodex}

\usepackage{placeins}

\AtBeginEnvironment{algorithmic}{\setstretch{1.2}}

\usepackage{pgfplots}
\pgfplotsset{compat=1.18}
\usepgfplotslibrary{groupplots}
\usetikzlibrary{external}
\usetikzlibrary{shapes.geometric}
\usepgfplotslibrary{fillbetween}
\usetikzlibrary{arrows.meta,positioning}

\title{Discrete Diffusion Inference-Time Control with Nested Sequential Monte Carlo}

\author{Lohithsai Yadala Chanchu, Hany Abdulsamad, Christian A. Naesseth 
\\
\textit{University of Amsterdam}}

\iclrfinalcopy 
\begin{document}

\maketitle

\begin{abstract}
We study inference-time control for text generation in discrete diffusion language models, where the goal is to steer sampling toward sequence-level rewards without retraining. Prior work in this domain has focused on particle-based methods such as best-of-$n$ sampling and bootstrap sequential Monte Carlo, which may suffer from overoptimism and weight degeneracy, respectively. We address these limitations using \emph{nested} sequential Monte Carlo methods. We formulate nested SMC (NSMC) and fully-adapted nested SMC (FA-NSMC) for Feynman--Kac steering, identifying and correcting  errors in prior formulations that lead to biased final estimates. We evaluate these methods on toxicity and fluency steering tasks, showing that NSMC and FA-NSMC consistently outperform best-of-$n$ and bootstrap SMC.
\end{abstract}

\section{Introduction}

Diffusion-based generative models have achieved remarkable success across continuous modalities, producing state-of-the-art results in image synthesis \citep{song2021scorebased}, video generation \citep{ho2022video}, and protein design \citep{gruver2023protein}. While autoregressive (AR) models have long dominated the landscape of text generation, the diffusion paradigm has recently expanded to the discrete domain of natural language processing \citep{sahoo2024simple, ye2025dream7bdiffusionlarge, shi2024simplified}, offering a compelling alternative.
Unlike standard autoregressive models that generate text token by token in a fixed left-to-right order, discrete diffusion language models (DDLMs) \citep{austin2021structured, sahoo2024simple,shi2024simplified} generate data through an iterative denoising process. Models such as the masked diffusion language model (MDLM) \citep{sahoo2024simple} learn a reverse-time Markov chain that progressively refines a sequence from a maximally corrupted degenerate state into coherent text, enabling bidirectional context integration and allowing the model to attend to information from all positions simultaneously to produce more globally consistent outputs.

\begin{figure}[t]
    \centering
    \input{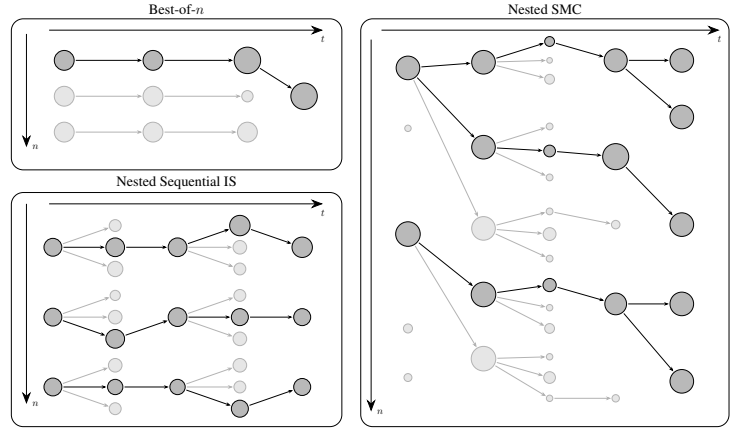}
    \caption{Illustration of different particle-based steering strategies.
    (Left) Best-of-$n$: independent proposals with selection based on exponentiated rewards.
    (Middle) Nested sequential importance sampling: particles propagate sequentially with weight updates but no resampling.
    (Right) Nested SMC: particle filtering approach where each outer particle spawns an inner SMC, approximating the locally optimal proposal distribution for improved effective sample size.}
    \label{fig:smc}
\end{figure}

Despite these architectural advantages, the capability to generate coherent text does not inherently ensure alignment with human intent or safety standards. In practice, we aim to generate samples that optimize specific downstream objectives, such as minimizing toxicity, while preserving the diversity and naturalness of the pre-trained model. Relying solely on the base model is often insufficient, as pre-trained models may reproduce undesirable biases found in their training data. Furthermore, while training-time alignment methods like reinforcement learning from human feedback (RLHF) \citep{ouyang2022training} are effective, they are computationally intensive, prone to mode collapse \citep{kirk2024understanding}, and rigidly couple the model to a single reward function. This motivates inference-time steering mechanisms that can flexibly guide discrete diffusion models toward user-specified rewards without the overhead of retraining. Broadly, existing approaches fall into two families:
\begin{itemize}[leftmargin=2em]
    \item Gradient-based methods, such as classifier guidance \citep{dhariwal2021}, modify the denoising drift using gradient information. These methods rely on differentiable reward functions, which significantly limits their applicability in discrete domains.
    \item Gradient-free methods, including best-of-$n$ sampling, rejection sampling \citep{na2024diffusionrejectionsampling}, and particle-based rare-event simulations \citep{naesseth2019smc, uehara2025inferencetime, li2024derivativefree, singhal2025framework}, do not require differentiability and therefore apply more generally, though they are often computationally intensive.
\end{itemize}
These limitations motivate the need for more efficient steering mechanisms that can flexibly guide discrete diffusion models toward arbitrary user-specified rewards. A promising framework is sequential Monte Carlo (SMC) \citep{naesseth2019smc, chopin2020introduction}, first studied for continuous diffusion models by \citet{wu2023practical,trippe2023diffusion,cardoso2024monte,dou2024diffusion}. SMC is a family of flexible probabilistic algorithms used to sample from complex sequences of distributions. At a high level, SMC maintains a population of \emph{particles}, each representing a potential partial text generation trajectory, that evolves over time. Through a process of mutation (proposing new tokens) and selection (reweighting and resampling based on the reward), SMC methods steer the population toward a modified version of the model's original distribution that favors desirable properties encoded by the reward function.

The quality of SMC samples depends critically on the proposal distribution, the mechanism used to mutate particles across time steps. In inference-time steering, the central challenge is that poor proposals are unlikely to generate samples associated with high-reward regions of the discrete text space, causing most particles to accrue low weights and degenerate rapidly. This leads to wasted computation and ineffective steering \citep{naesseth2019smc}. Recent attempts to adapt SMC to discrete diffusion models face persistent difficulties stemming from proposal design.


\citet{singhal2025framework} formulate Feynman–Kac (FK) steering, and use bootstrap proposals in practice where new particle candidates are generated using the pretrained base model. While straightforward to implement, this proposal is agnostic to the reward, leaving undesirable particles to be filtered only through subsequent reweighting. This method often exhibits low statistical efficiency and struggles to discover rare, high-reward paths. Soft value-based decoding (SVDD) \citep{li2024derivativefree} takes a different approach by casting steering as nested sequential importance sampling (SIS). However, this method inherits the well-known pathologies of (nested) SIS methods, including weight degeneracy and high variance over long horizons \citep{naesseth2019smc}. Alternatively, \citet{taylorseries26} construct improved proposals by leveraging gradient information of the reward function. In discrete text domains, this typically requires continuous relaxations, which may introduce approximation error.

To address these limitations, we propose leveraging nested SMC (NSMC) methods \citep{naesseth2015nsmc, naesseth2019high}, which introduce an internal SMC sampler to approximate the locally optimal proposal and the associated normalizing constants when these quantities are not available in closed form. NSMC runs an inner SMC procedure for each outer particle to estimate the optimal proposal; the inner sampler produces (i) a properly weighted sample used to draw the child state, and (ii) an unbiased Monte Carlo estimate of the predictive normalizing constant required for the correct outer weight update. The fully-adapted NSMC (FA-NSMC) method further refines this idea by using estimated predictive weights to resample parents before propagation, increasing particle diversity.

Although NSMC is well-established in computational statistics, it has not been applied to steering in modern discrete diffusion language models. A recent tutorial on diffusion-guidance by \citet{uehara2025inferencetime} which, building on \citet{li2024derivativefree}, presents an algorithm labeled ``nested SMC''. However, its weighting scheme does not correspond to a properly weighted NSMC algorithm, leading to systematic bias even in the infinite-particle limit. We resolve this by developing correctly weighted NSMC variants, implementing them for a discrete diffusion language model, and evaluating them on toxicity and perplexity steering tasks, which provide controlled environments for understanding how SMC variants behave in practice.

We summarize our contributions as follows:
\begin{itemize}
    \item We develop properly weighted nested SMC and fully-adapted NSMC updates for Feynman--Kac steering in discrete diffusion language models.
    \item We empirically compare NSMC and FA-NSMC against bootstrap SMC baselines on toxicity and perplexity steering tasks, characterizing when nested methods improve sample efficiency and controllability.
\end{itemize}

\section{Background}

We start by introducing the notation for discrete diffusions, the tilted path measures that correspond to the aligned sampling targets, the corresponding Feynman--Kac model, and the SMC algorithm.

\paragraph{Diffusion Models.} Let $\mathcal{V}$ be a finite vocabulary of tokens and $\mathcal{X} = \mathcal{V}^{L}$ be the state space of a sequence of length $L$. We consider a diffusion process over token sequences, discretized into $T + 1$ time steps $t \in \{0, \dots, T\}$. Here, $t=T$ represents the maximally corrupted, fully masked state, and $t=0$ represents the clean generated sequence. Given a pre-trained reverse-time generative base model, the prior path measure over trajectories $x_{0:T} \coloneq (x_{0}, \dots, x_{T}) \in \mathcal{X}^{T+1}$, conditioned on a context or prompt $c$, factorizes as:
\begin{equation}
    \label{eq:prior-path}
    p(x_{0:T} \mid c) = \mu(x_{T})\prod_{t=1}^{T} f(x_{t-1} \mid x_{t}, c),
\end{equation}
where $\mu(\cdot)$ is the fixed distribution at $t=T$ of fully masked sequences and $f(\cdot \mid x_{t}, c)$ denotes the reverse transition kernel used to denoise the sequence from step $t$ to $t-1$. For notational simplicity, we omit the dependence on $c$ hereafter and write $p(x_{0:T})$ and $f(\cdot \mid x_t)$.

\paragraph{Target Distribution.} We want to sample from a distribution aligned with a scalar reward $r:\!\mathcal{X}\to\mathbb{R}$ evaluated on the terminal state $x_0$. For $\lambda > 0$, we define the \emph{reward-tilted} terminal distribution 
\begin{equation}
    p_{\lambda}(x_{0}) = \frac{p(x_{0}) \exp \left( \lambda \, r(x_{0}) \right)}{ Z_{\lambda}}, \quad \text{where} \quad Z_{\lambda} \coloneq \mathbb{E}_{x_{0} \sim p} \big[ \exp \left( \lambda \, r(x_{0}) \right) \big].
\end{equation}
To sample from $p_{\lambda}(x_{0})$, we define the unnormalized target path measure $\gamma_{0}(x_{0:T})$ by weighting the prior path measure by the terminal reward:
\begin{equation}
    \label{eq:tilted-path}
    \gamma_{0}(x_{0:T}) = \left\{ \mu(x_{T}) \prod_{t=1}^{T} f(x_{t-1} \mid x_{t}) \right\} \exp \left( \lambda \, r(x_{0}) \right).
\end{equation}

\paragraph{Feynman--Kac.} We frame the problem of sampling from $\gamma_{0}(x_{0:T})$ in terms of a Feynman--Kac model~\citep{del2004feynman}. A FK model is characterized by the transition kernel $f(\cdot \mid x_{t})$ and a set of nonnegative potential functions $G_{t-1}\!: \mathcal{X} \times \mathcal{X} \to \mathbb{R}^{+}$. The induced path measure is:
\begin{equation}
    \pi(x_{0:T}) \propto \mu(x_{T}) \, G_{T}(x_{T}) \left\{ \prod_{t=1}^{T} f(x_{t-1} \mid x_{t}) \, G_{t-1}(x_{t-1}, x_{t}) \right\}.
\end{equation}
To recover our specific target $\gamma_{0}(x_{0:T})$, the potentials must telescope to reproduce the desired tilt, satisfying $G_{T}(x_{T}) \prod_{t=1}^{T} G_{t-1}(x_{t-1}, x_{t}) = \exp \left( \lambda \, r(x_{0}) \right)$. While one could set $G_{0}(\cdot) = \exp \left( \lambda \, r(x_{0}) \right)$ and $\left\{G_{t} \right\}_{t>0} = 1$, this choice yields an inefficient sampling procedure that suffers from path degeneracy and high-variance weights at $t=0$. Since intermediate potentials provide no signal with regard to high-reward regions, samples are propagated under the base dynamics and only receive their reward-dependent weighting at the last step.

\paragraph{Sequential Monte Carlo.} SMC \citep{naesseth2019smc,chopin2020introduction} is a sampling method designed to approximate a sequence of intermediate unnormalized targets $\gamma_{t}(x_{t:T})$, with corresponding normalized targets $\pi_t\propto \gamma_t$. To approximate $\pi_t$, SMC uses a set of weighted samples, or \emph{particles}, $\{(w_t^{(i)},x_{t:T}^{(i)})\}_{i=1}^N$,
\begin{align*}
    \pi_t(x_{t:T}) &\approx \sum_{i=1}^N w_t^{(i)}\delta_{x_{t:T}^{(i)}},
\end{align*}
where $\delta_X$ is the Dirac measure at $X$. The particle system is then updated from time $t$ to $t-1$ by repeating the following for each particle $i$: 
\begin{enumerate}
    \item Resampling, $a\sim \mathrm{Cat}\left(w_t^{1:N}\right)$, 
    \item Propagation, $x_{t-1}^{(i)} \sim q_{t-1}(x_{t-1}|x_{t}^{(a)})$,
    \item Weighting, $w_{t-1}^{(i)} \propto \frac{\displaystyle \gamma_{t-1}((x_{t-1}^{(i)},x_{t:T}^{(a)}))}{\displaystyle \gamma_{t}(x_{t:T}^{(a)}) q_{t-1}(x_{t-1}^{(i)}|x_{t}^{(a)})}$.
\end{enumerate}
The key design variables are the intermediate targets $\gamma_t$ and the proposals $q_{t}$. \emph{Bootstrap} SMC for the FK model in Algorithm~\ref{alg:smc} is obtained by setting $q_{t-1}(x_{t-1}|x_t)=f(x_{t-1}|x_t)$ and $\gamma_{t-1}(x_{t-1:T}) = \gamma_t(x_{t:T}) f(x_{t-1}|x_t) G_{t-1}(x_{t-1},x_t)$.

\paragraph{Optimal Twisting.} To provide intermediate guidance, we construct the targets $\gamma_t$ by \emph{twisting} the prior path measure $p(x_{t:T})$ with a set of positive potential functions $\psi_{t}\!: \mathcal{X} \to \mathbb{R}^{+}$ that look ahead and tilt the intermediate targets toward high-reward regions $\gamma_{t}(x_{t:T}) \coloneq p(x_{t:T}) \, \psi_{t}(x_{t})$. \citet{naesseth2019smc,whiteley2014twisted,guarniero2017iterated,heng2020controlled} identify the optimal twisting functions, which minimize the variance of the incremental weights, as the conditional expectation of the future reward:
\begin{equation}
    \label{eq:optimal-twisting}
    \psi_{t}^{\star}(x_t) \coloneq \mathbb{E}_{p} \big[\exp \left(\lambda \, r(x_0) \right)\mid x_{t}\big], \quad t=0,\dots,T.
\end{equation}
At the terminal step $t = 0$, this definition recovers the exact reward tilt $\psi^{\star}_{0}(x_{0}) = \exp \left( \lambda \, r(x_{0}) \right)$.

\paragraph{Optimal Proposals.} The optimal twisting functions naturally induce a sequence of corresponding optimal proposal kernels that realize the transition between the intermediate targets:
\begin{equation}
    \label{eq:optimal-proposal}
    q_{t-1}^\star(x_{t-1} \mid x_{t}) \propto \frac{\gamma_{t-1}(x_{t-1:T})}{\gamma_{t-1}(x_{t:T})} = \frac{f(x_{t-1} \mid x_{t}) \, \psi_{t-1}^{\star}(x_{t-1})}{\psi_{t}^{\star}(x_{t})}, \qquad t=1,\dots,T,
\end{equation}
while at time $t=T$, the optimal proposal is given by: $q^{\star}_{T}(x_{T}) \propto \mu(x_{T})\,\psi_{T}^{\star}(x_{T})$. This in turn allows us to identify the optimal potential functions $\left\{G^{\star}_{t}\right\}_{t\geq0}$ as the ratios of successive twists:
\begin{equation}
    \label{eq:optimal-potential}
    G^{\star}_{t-1}(x_{t-1}, x_t) = \psi^{\star}_{t-1}(x_{t-1}) / \psi^{\star}_{t}(x_{t}), \qquad G^{\star}_T(x_T) = \psi^{\star}_T(x_T).
\end{equation}
Appendices~\ref{app:potential} and~\ref{app:telescoping} provide details and show that the cumulative product of these optimal potentials telescopes to recover the required terminal reward tilt. In contrast, the ``nested SMC'' algorithm in \citet{uehara2025inferencetime} uses only the numerator $\psi_{t-1}^\star(x_{t-1})$ in its weighting scheme, omitting the normalization by $\psi_{t}^\star(x_{t})$ implied by $G_{t-1}^{\star}$, and therefore fails to target the correct tilted distribution.

\section{Nested Sequential Monte Carlo}
Nested sequential Monte Carlo (NSMC) \citep{naesseth2015nsmc,naesseth2019high} is a class of particle algorithms that lets us derive practical algorithms for optimal twisting and proposal distributions.

First, recall that the optimal proposal is given by:
\begin{equation}
    q^{\star}_{t-1}(x_{t-1} \mid x_{t})= \frac{1}{\nu_{t-1}(x_{t})} \frac{\gamma_{t-1}(x_{t-1:T})}{\gamma_{t}(x_{t:T})} \propto \frac{\gamma_{t-1}(x_{t-1:T})}{\gamma_{t-1}(x_{t:T})} 
\end{equation}
where $\nu_{t-1}(x_{t})$ is the predictive normalizing constant:
\begin{equation}
    \label{eq:predictive-normalizer}
    \nu_{t-1}(x_{t}) \coloneq \sum_{x_{t-1} \in \mathcal{V}^L} \!\! \frac{\gamma_{t-1}(x_{t-1:T})}{\gamma_{t}(x_{t:T})} = \mathbb{E}_{x_{t-1} \sim f(\cdot \mid x_{t})} \Big[G^{\star}_{t-1}(x_{t-1}, x_{t}) \Big].
\end{equation}
Under $q_{t-1}^\star$, the incremental importance weights $w_{t-1}$ are equal to this normalizing constant: $w_{t-1} = \gamma_{t-1}(x_{t-1:T}) / (q^{\star}_{t-1}(x_{t-1} \mid x_{t}) \, \gamma_{t}(x_{t:T})) \equiv \nu_{t-1}(x_{t})$. This means it does not depend on the particular sample of $x_{t-1}$. This is the key variance-reduction property of optimal proposals: conditional on the parents $x_{t}$, the incremental weights $w_{t-1}$ are uniform and their incremental variance is zero.

Computing the predictive normalizer $\nu_{t-1}(x_{t})$ is generally intractable in high-dimensional spaces, as it requires summing over all $\left| \mathcal{V} \right|^{L}$ possible sequences at every step. NSMC resolves this intractability by replacing $\nu_{t-1}(x_{t})$ with an inner Monte Carlo estimate.

For each parent particle $x_{t}^{(i)}$, where $i \in \{1, \dots, N\}$, we propose $M$ candidate states from the base transition kernel $x_{t-1}^{(i,j)} \sim f(\cdot \mid x_{t}^{(i)})$, for $j=1,\dots,M$. For each candidate, we compute an inner importance weight $v_{t-1}^{(i,j)}$ by evaluating the optimal twisting potential functions:
\begin{equation}
    v_{t-1}^{(i,j)} \coloneq \frac{\gamma_{t-1}(x_{t-1:T}^{(i,j)})}{\gamma_{t}(x_{t:T}^{(i)}) \, f(x_{t-1}^{(i,j)} \mid x_{t}^{(i)})} = G^{\star}_{t-1}(x_{t-1}^{(i,j)}, x_{t}^{(i)})
\end{equation}
where $x_{t-1:T}^{(i,j)}=(x_{t-1}^{(i,j)}, x_{t:T}^{(i)})$. The predictive normalizing constant is then approximated by the average of the inner weights $\hat{\nu}_{t-1}(x_{t}^{(i)}) \coloneq 1/M \sum_{j=1}^{M} v_{t-1}^{(i,j)} \approx \nu_{t-1}(x_{t}^{(i)})$. 

To approximate sampling from the optimal proposal $q^{\star}(\cdot \mid x_{t})$, NSMC selects a single candidate trajectory for the next step by resampling from the candidates based on their inner weights:
\begin{equation}
    b^{(i)} \sim \mathrm{Cat} \left( \left\{ \frac{v_{t-1}^{(i,j)}}{\sum_{k} v_{t-1}^{(i,k)}} \right\}_{j=1}^{M} \right),
\end{equation}
and setting $x_{t-1}^{(i)} \gets x_{t-1}^{(i,\ell)}$, where $\ell = b^{(i)}$. This nested approach ensures that the outer incremental importance weights $w_{t-1}^{(i)} = \hat{\nu}_{t-1}(x_{t}^{(i)})$ remain unbiased estimates of the true normalizing constants. Algorithm~\ref{alg:nested-proposal} provides an overview of the nested proposal procedure.

By contrast, a standard bootstrap SMC, Algorithm~\ref{alg:smc}, uses a single candidate proposal per parent without incorporating reward information, leading to resampling decisions based on a noisy one-sample estimate of future potential. NSMC reduces this noise by averaging over $M$ candidates to estimate the predictive normalizer $\nu_{t-1}$ and uses the inner weights to bias candidate selection toward promising regions of future high reward. Algorithm~\ref{alg:nsmc} provides a detailed recipe for NSMC.

Finally, the \emph{fully-adapted} NSMC (FA-NSMC) procedure incorporates lookahead information into the parent resampling mechanism. In this scheme, we estimate the future potential of all particles before committing to the resampling step. We perform the same lookahead procedure as in NSMC to generate $M$ candidates and estimate the predictive normalizer for every parent $\hat{\nu}_{t-1}(x_{t}^{(i)})$.

Unlike standard NSMC, which resamples parents based solely on their accumulated weights $w_{t}^{(i)}$, FA-NSMC resamples parent indices $a^{(i)}$ proportional to $w_{t}^{(i)} \cdot \hat{\nu}_{t-1}(x_{t}^{(i)})$. Once the parent $k = a^{(i)}$ is selected, we sample $x_{t-1}^{(i)}$ from that parent's candidates using the inner weights $v_{t-1}^{(k,\cdot)}$. 

Crucially, this reordering enhances sample diversity. If a high-potential parent $x_{t}^{(i)}$ is selected multiple times, we can draw multiple \emph{distinct} children from its set of promising candidates. In contrast, the standard resampling scheme would simply replicate the \emph{same} parent state multiple times. This fully-adapted procedure is described in Algorithm~\ref{alg:fa-nsmc}.

\begin{algorithm}[t]
    \caption{Nested Proposal}
    \label{alg:nested-proposal}
    \begin{algorithmic}
    \Procedure{NestedProposal\,}{$x_{t}, f, G^\star$}
        \State Sample candidates $x_{t-1}^{(j)}\sim f(\cdot \mid x_{t})$, for $j=1,\dots,M$
        \State Compute weights $v^{(j)} \gets G_{t-1}^\star(x_{t-1}^{(j)}, x_{t})$
        \State Estimate normalizer $\hat\nu \gets 1/M \sum_{j=1}^M v^{(j)}$
        \State \Return $x_{t-1}^{1:M}$, $v^{1:M}$, $\hat\nu$
    \EndProcedure
    \end{algorithmic}
\end{algorithm}

\begin{algorithm}[t]
    \caption{Nested SMC}
    \label{alg:nsmc}
    \begin{algorithmic}
    \Require Base kernel $f$, potentials $\left\{ G^{\star}_{t} \right\}_{t\geq0}$, population sizes $N, M$
    \State Set $w_{T}^{(i)} \gets 1/N$, for $i=1,\dots,N$
    \State Sample $x_{T}^{(i)} \sim \mu(x_T) \, G^{\star}_{T}(x_{T})$
    \For{$t=T$ \textbf{to} $1$}
        \For{$i=1$ \textbf{to} $N$}
            \State Resample $a \sim \mathrm{Cat}(w_{t}^{1:N})$
            \State $x_{t-1}^{(i,\cdot)}, v^{(i,\cdot)}, \hat\nu^{(i)} \gets$ \Call{NestedProposal}{$x_t^{(a)}, f, G^{\star}_{t-1}$}
            \State Sample $b \sim \mathrm{Cat}\left(v^{(i,\cdot)} / \sum_k v^{(i,k)}\right)$
            \State Set $x_{t-1}^{(i)} \gets x_{t-1}^{(i,b)}$, $\tilde w_{t-1}^{(i)} \gets \hat\nu^{(i)}$
        \EndFor
        \State Normalize $w_{t-1}^{(i)} \gets \tilde w_{t-1}^{(i)} / \sum_{k=1}^N \tilde w_{t-1}^{(k)}$
    \EndFor
    \State \Return $\{(x_0^{(i)}, w_0^{(i)})\}_{i=1}^N$
    \end{algorithmic}
\end{algorithm}

\section{Numerical Evaluation}

We evaluate the performance of several inference-time sampling algorithms for reward-tilted generation within the framework of discrete diffusion language models. Specifically, we benchmark our proposed nested SMC (NSMC) and fully-adapted nested SMC (FA-NSMC) against two baselines: Best-of-$N$ (BoN) and Bootstrap SMC. Beyond measuring the terminal reward, we analyze: 
(i) the effect of steering on longer texts,
(ii) the influence of population sizes $N$ and $M$,
(iii) intermediate reward evolution under the weighting scheme presented in \citet{uehara2025inferencetime}, and
(iv) the impact of the number of samples $K$ used to approximate the optimal potentials.

\subsection{Experimental Setup}

All algorithms above assume access to a terminal reward $r(x_0)$ and the ideal twisting functions and potentials defined in~\eqref{eq:optimal-twisting} and~\eqref{eq:optimal-potential}. In practice, we approximate the conditional expectations with a tractable surrogate, estimating the future reward using the model's single-step prediction $\hat{x}_0$ at each state. The particle-based algorithms themselves are unchanged, and the theoretical guarantees for $t=0$ remain valid.
 
Our experimental validation focuses on two distinct steering objectives:
\begin{itemize}[leftmargin=2em]
    \item \emph{Toxicity steering:} We define $r(x_0) = r_{\text{tox}}(x_0)$ using a toxicity classifier, encouraging the generation of toxic content to test alignment.
    \item \emph{Fluency steering:} We define $r(x_0) = -r_{\text{ppl}}(x_0)$, penalizing high perplexity to encourage the generation of fluent text.
\end{itemize}
We maintain a consistent algorithmic framework across these tasks, varying only the scalar reward function used to define the exponential tilt $\exp(\lambda r(x_0))$.

\textbf{Base Model.} We steer the publicly released MDLM \citep{sahoo2024simple} discrete diffusion model checkpoint, a DiT-style architecture with $12$ transformer blocks, $12$ attention heads, and $768$ hidden units trained on OpenWebText with a GPT-2 tokenizer. Unless otherwise stated, generations are produced using $T=50$ diffusion steps. Following \citet{han2023}, we use $15$ controllable-generation prompts (e.g., ``Once upon a time'', ``The book'', ``The year is 1910.''). For each prompt and configuration, we sample $10$ independent continuations and report metrics averaged over the resulting $15\times 10 = 150$ generations.

\textbf{Generation Protocol and Resampling Schedule.}
We generate 100-token continuations and resample at every diffusion step, unless stated otherwise. For ablations that vary the reward window length, we generate sufficiently long continuations, such that the reward window is well-defined.

\textbf{Toxicity Reward.} 
For toxicity steering, we use an off-the-shelf RoBERTa toxicity classifier\footnote{\url{https://huggingface.co/s-nlp/roberta_toxicity_classifier}} \citep{logacheva} and define $r_{\text{tox}}(x_0)$ as the log-softmax score of the toxic class (no clipping/normalization). We fix the steering strength parameter $\lambda=10$ for all toxicity experiments. During steering, rewards are computed on a fixed continuation suffix length (reward window). For evaluation, we concatenate the full prompt and continuation and additionally report toxicity rates under a holdout classifier\footnote{\url{https://huggingface.co/textdetox/xlmr-large-toxicity-classifier}} \citep{Dementieva}, to assess robustness. 

\textbf{Perplexity Reward.}
For perplexity-based rewards (used only in the perplexity-steering task), we use GPT-2-XL \citep{Radford2019LanguageMA} to score intermediate $x_0$ reconstructions.

\textbf{Intermediate Potentials via $x_0$ Reconstructions.}
The described particle methods require intermediate potentials that approximate the remaining terminal reward. At resampling time $t$, for each particle state $x_t^{(i)}$, we draw $K$ samples $\{\hat x_0^{(i,k)}\}_{k=1}^K$ and form the estimator
\begin{equation}
    \widehat{\psi}_t(x_t^{(i)}) = \frac{1}{K}\sum_{k=1}^K \exp \big(\lambda\, r(\hat x_0^{(i,k)})\big),    
\end{equation}
which enters the importance weights at resampling. We compare $K\in\{4,16\}$ to ablate the effect of the reconstruction count. We also log $\widehat{\psi}_t$ over $t$ to study how reward information propagates along the reverse chain. Additional plots are provided in Appendix~\ref{inter-plots}. Replacing the true potential with an unbiased estimator still results in a properly weighted SMC algorithm \citep[Section 4.3]{naesseth2019smc}.

\textbf{Compute Budgets.}
We match compute by the number of forward passes per diffusion step. With $N$ outer particles, bootstrap SMC, NSMC, and FA-NSMC each require $N$ diffusion-model evaluations per step. For (FA-)NSMC, the $M$ inner proposals are drawn by categorical sampling from already-computed logits (no additional transformer evaluations). Reward model calls are also typically cheaper than diffusion forward passes, so we treat $N$ as the primary hyperparameter and vary $(N,M,K)$ under this constraint. For BoN, we match compute by setting $N$ so that the total number of diffusion-model evaluations matches the particle methods, similar to \citet{singhal2025framework}.

\textbf{Metrics.}
We report: (i) toxicity rates under a binary toxicity classifier and a separate holdout binary toxicity classifier, (ii) perplexity as a fluency proxy, and (iii) output diversity via Distinct-1/Distinct-2 \citep{tevet-berant-2021-evaluating} (see Appendix~\ref{tab:diversity}). Perplexity is not optimized in toxicity steering.

\subsection{Results and Discussion}



\begin{table}[t]
\centering
\begin{minipage}[t]{0.48\textwidth}
\centering
\caption{Toxicity and fluency steering ($N=4$, $M=8$, $K=4$, $\lambda=10$). PPL via GPT-2-XL.}
\label{tab:main_toxicity_gpt2}
\small
\begin{tabular}{lcc}
\toprule
Method & Toxic $\uparrow$ & PPL $\downarrow$ \\
\midrule
Base (MDLM) & 0.003 & 85.3 \\
BoN & 0.022 & 55.5 \\
SMC (bootstrap) & 0.25 & 49.0 \\
NSMC & 0.39 & \textbf{42.3} \\
FA-NSMC & \textbf{0.40} & 42.9 \\
\bottomrule
\end{tabular}
\end{minipage}
\hfill
\begin{minipage}[t]{0.48\textwidth}
\centering
\caption{Effect of reward window length on toxicity rate ($N=8$, $M=8$, $K=4$).}
\label{tab:reward_length_toxicity}
\small
\begin{tabular}{c|ccc}
\toprule
Length & SMC & NSMC & FA-NSMC \\
\midrule
50  & \textbf{0.57} & \textbf{0.70} & \textbf{0.68} \\
100 & 0.40 & 0.50 & 0.51 \\
300 & 0.29 & 0.30 & 0.47 \\
\bottomrule
\end{tabular}
\end{minipage}
\end{table}



\begin{table}[t]
\centering
\begin{minipage}[t]{0.32\textwidth}
\centering
\caption{Effect of population size $M$ ($N{=}8$, $K{=}4$).}
\label{tab:inner_sweep}
\scriptsize
\begin{tabular}{lccc}
\toprule
Method & $M$ & Tox $\uparrow$ & Hold $\uparrow$ \\
\midrule
NSMC & 1  & .57 & .48 \\
NSMC & 2  & .61 & .44 \\
NSMC & 4  & .59 & .56 \\
NSMC & 8  & .71 & .56 \\
NSMC & 16 & \textbf{.71} & \textbf{.62} \\
NSMC & 32 & .70 & .55 \\
\midrule
FA-NSMC & 1  & .54 & .47 \\
FA-NSMC & 2  & .58 & .49 \\
FA-NSMC & 4  & .62 & .55 \\
FA-NSMC & 8  & .68 & .60 \\
FA-NSMC & 16 & \textbf{.74} & \textbf{.66} \\
FA-NSMC & 32 & .71 & .51 \\
\bottomrule
\end{tabular}
\end{minipage}
\hfill
\begin{minipage}[t]{0.65\textwidth}
\centering
\caption{Results over 10 repititions ($\lambda{=}10$).}
\label{tab:tox_repeated}
\scriptsize
\begin{tabular}{c|lc|ccc||ccc}
\toprule
& & & \multicolumn{3}{c||}{$K=4$} & \multicolumn{3}{c}{$K=16$} \\
\cline{4-9}
$N$ & Method & $M$ & \rule{0pt}{2.4ex} Tox $\uparrow$ & Hold $\uparrow$ & PPL $\downarrow$ & Tox $\uparrow$ & Hold $\uparrow$ & PPL $\downarrow$ \\
\midrule
\multirow{5}{*}{4}
& SMC & -- & .25 & .19 & 49 & .31 & .36 & 44 \\
& NSMC & 4  & .29 & .21 & 39 & .46 & .44 & 47 \\
& NSMC & 8  & .39 & .31 & 42 & .45 & .40 & 42 \\
& FA-NSMC & 4  & .36 & .38 & \textbf{38} & \textbf{.48} & \textbf{.44} & 41 \\
& FA-NSMC & 8  & \textbf{.40} & \textbf{.39} & 43 & .45 & .43 & \textbf{40} \\
\midrule
\multirow{5}{*}{8}
& SMC & -- & .57 & .48 & 47 & .67 & .59 & 41 \\
& NSMC & 4  & .70 & .56 & 36 & .71 & .64 & 36 \\
& NSMC & 8  & \textbf{.70} & .56 & 38 & .74 & .59 & 38 \\
& FA-NSMC & 4  & .62 & .55 & \textbf{33} & .70 & \textbf{.61} & 37 \\
& FA-NSMC & 8  & .68 & \textbf{.60} & 39 & \textbf{.74} & .59 & \textbf{32} \\
\bottomrule
\end{tabular}
\end{minipage}
\end{table}


\textbf{Steering Results for Toxicity and Fluency Tasks (Table~\ref{tab:main_toxicity_gpt2}).}
Table~\ref{tab:main_toxicity_gpt2} compares BoN, bootstrap SMC, NSMC, and FA-NSMC. Nested methods substantially improve the toxicity rate over both BoN and bootstrap SMC, with FA-NSMC slightly outperforming NSMC. The base MDLM rarely produces toxic continuations, reflecting the rarity of toxicity under the base model. Best-of-$n$ yields only a marginal increase, because it selects from a small set of fully sampled $x_0$ candidates offering limited leverage when high-reward outcomes are rare. In contrast, bootstrap SMC achieves a much larger increase by reallocating computation toward partial trajectories whose intermediate reconstructions already score highly under the reward.

Nested methods improve outcomes by reducing the discrepancy between the proposal and the reward-tilted target. Intermediate potentials provide a lookahead estimate of future reward, yielding more informative resampling and reduced weight degeneracy. In this configuration, the difference between NSMC and FA-NSMC is small. The effect is more pronounced for perplexity steering, where nested methods show a larger gain over bootstrap SMC relative to BoN.

\textbf{Reward Window Length Sensitivity (Table~\ref{tab:reward_length_toxicity}).}
Table~\ref{tab:reward_length_toxicity} sweeps the length of the continuation suffix used by the reward model at fixed $N=8$ and $M=8$. As the reward window grows, performance degrades across methods. This is expected: longer suffixes make toxicity rarer and noisier to predict, increase the chance of drifting away from toxic content, and introduce greater long-horizon uncertainty early in the reverse process. As a result, intermediate potentials become less informative---reconstruction-based reward estimates have higher variance and are less predictive of the terminal reward---reducing the effectiveness of resampling and increasing particle impoverishment.

The notable exception is FA-NSMC at reward length 300. Full adaptation is most beneficial when lookahead is hardest: with long reward windows, accounting for future reward contributions at the proposal stage is more effective than relying on noisy weight corrections. When reward information is strongly delayed, better adaptation yields larger gains.

\begin{figure}[t]
    \centering
    \begin{subfigure}[t]{0.47\textwidth}
        \centering
        \includegraphics[width=\linewidth]{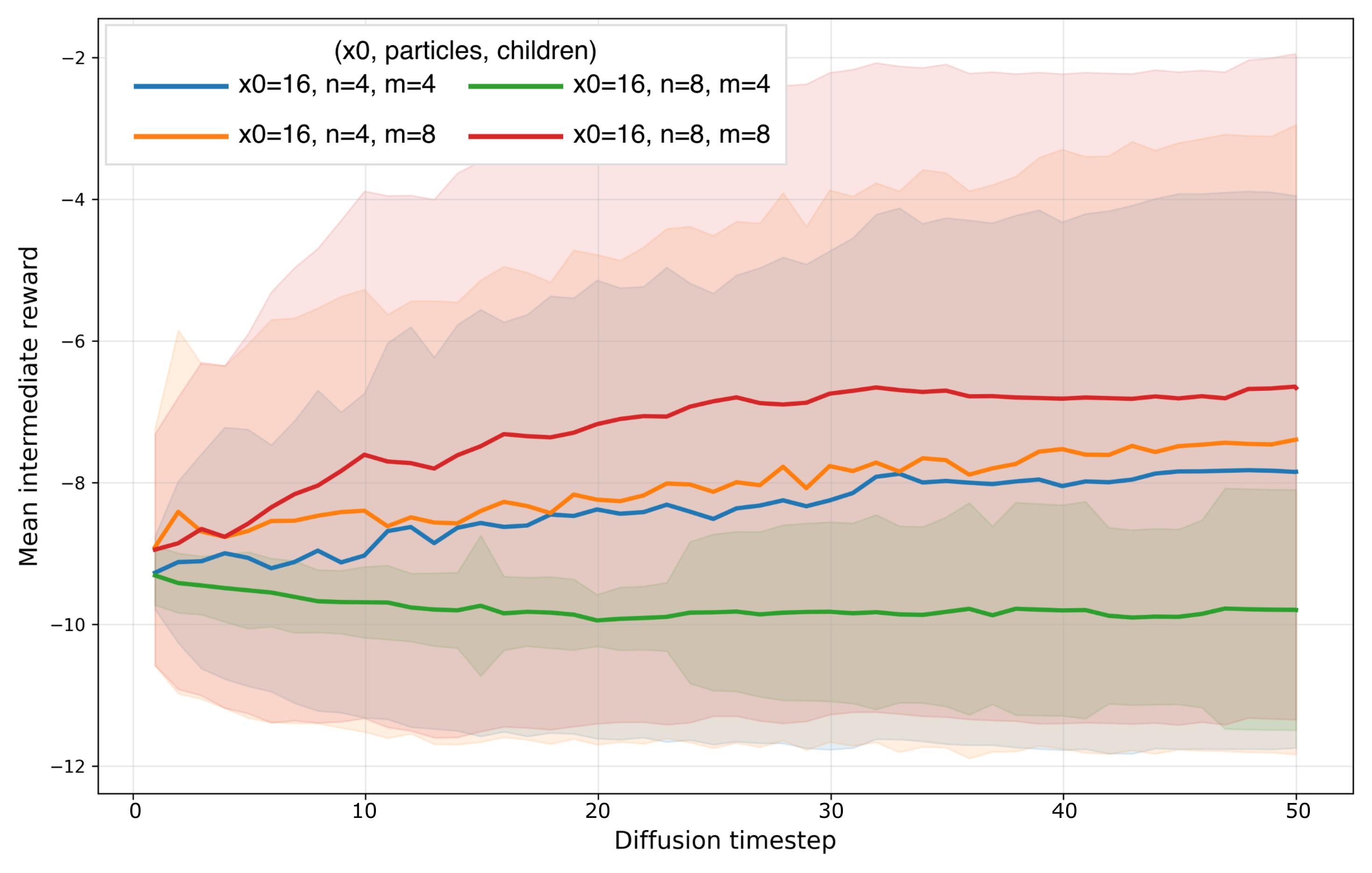}
        \caption{Average intermediate toxic reward for the biased NSMC variant in \citet{uehara2025inferencetime}.}
        \label{fig:flat-inter}
    \end{subfigure}
    \hfill
    \begin{subfigure}[t]{0.5\textwidth}
        \centering
        \includegraphics[width=\linewidth]{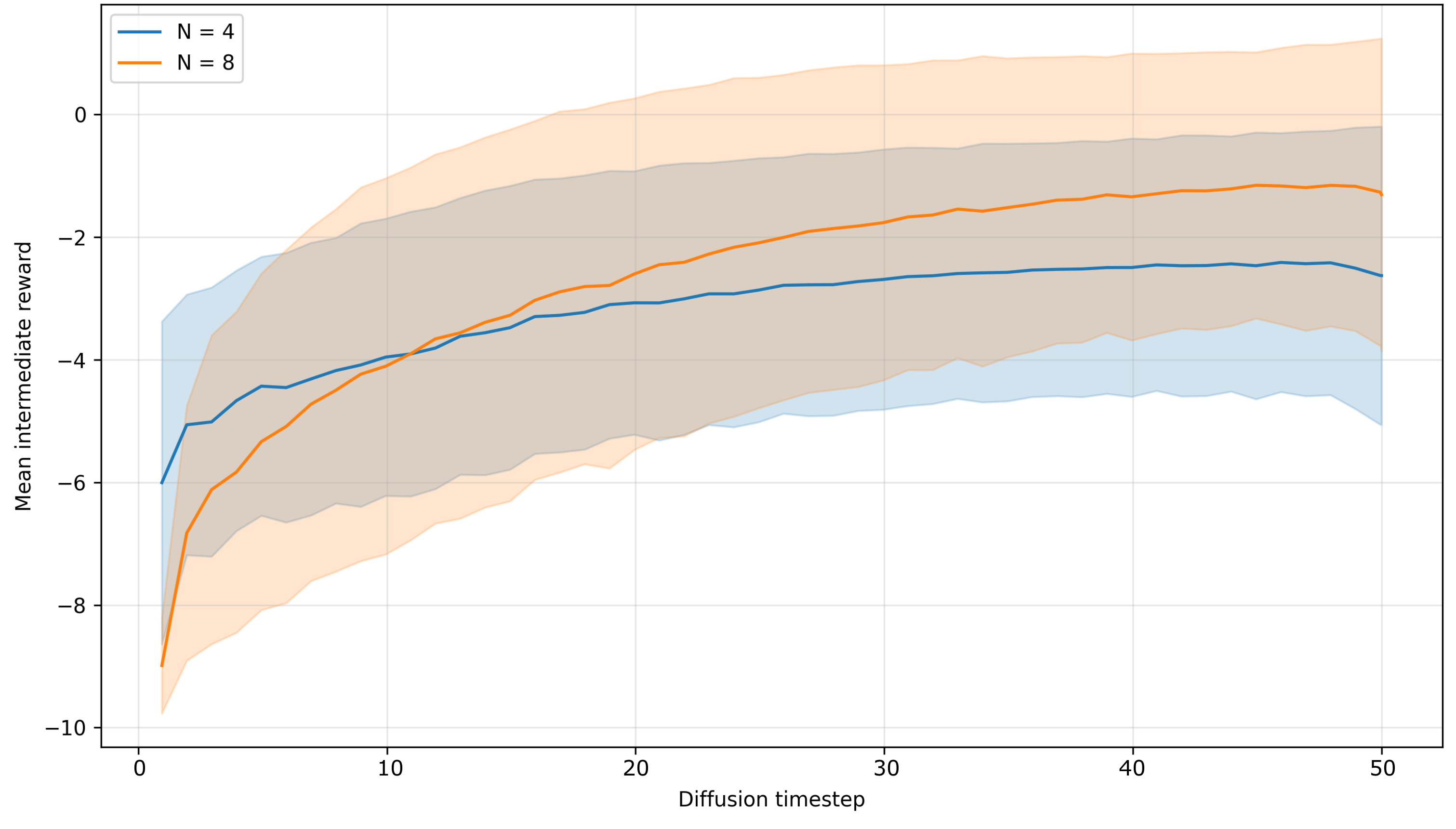}
        \caption{Average intermediate toxic reward with $N$ outer particles according to Algorithm \ref{alg:nsmc}}.
        \label{fig:intermediate_rewards_by_N}
    \end{subfigure}
    
    \caption{Comparison of intermediate rewards for toxicity steering.}
    \label{fig:combined_rewards}
\end{figure}

\textbf{Intermediate NSMC Rewards with Biased Potential (Figure~\ref{fig:combined_rewards}).}
We notice that intermediate expected rewards, $\mathbb{E}_p\left[\left(r(x_0)\right)|x_t\right]$, improve over time for NSMC and FA-NSMC when using the correct potentials $G^{\star}_{t-1}(x_{t-1}, x_{t})$, as shown in Figure~\ref{fig:intermediate_rewards_by_N}. In contrast, Figure~\ref{fig:flat-inter} shows that the potential proposed by \citet{uehara2025inferencetime}, which omits the denominator term, fails to target the correct distribution $p_\lambda(x_0) \propto p_\theta(x_0)\exp(\lambda r(x_0))$. As a result, rewards do not increase over time under this biased potential. A full sweep of toxicity rates along $N,M,K$ with \citet{uehara2025inferencetime}'s implementation is found in Appendix~\ref{app:wrong_nested}.

\textbf{Scaling with $(N,M,K)$ and Robustness (Tables~\ref{tab:inner_sweep} and \ref{tab:tox_repeated}).}
Table~\ref{tab:inner_sweep} sweeps $M$ for NSMC and FA-NSMC at fixed $N=8$, $K=4$.
Table~\ref{tab:tox_repeated} reports toxicity metrics over a broader sweep, averaged over 10 repetitions. Three trends stand out.
First, the number of outer particles $N$ dominates performance: increasing $N$ yields the largest gains, reflecting reduced Monte Carlo error.
Second, increasing the number of reconstructions $K$ improves guidance, especially at small $N$. This is consistent with the role of $K$ in reducing the variance of $\widehat{\psi}_t$, which helps preserve high-reward trajectories early.
Third, increasing the number of inner proposals $M$ yields gains that saturate quickly, indicating diminishing returns once the proposal is ``good enough''. The external toxicity rates and perplexity provide a useful sanity check against reward-model overfitting.

\section{Conclusion and Limitations}

This work provides initial evidence that nested sequential Monte Carlo methods can improve
inference-time steering for discrete diffusion language models. We show that nested methods, including fully adapted variants, achieve higher rewards than bootstrap SMC at fixed $N$, highlighting the value of better proposals and more informative intermediate potentials. FA-NSMC is the most robust variant in the most challenging regimes: it degrades less as the reward window grows and often improves external toxicity at comparable internal toxicity, suggesting improved robustness to reward-model idiosyncrasies. Furthermore, our experiments show that compute allocation matters: the number of outer particles $N$ is the primary factor, while increasing the number of reconstructions $K$ and inner proposals $M$ provides additional gains with diminishing returns. 


Our evaluation has clear limitations. We study only two reward settings, toxicity and perplexity, on a single base checkpoint, leaving open the question of how consistently these gains transfer across models and domains. In addition, our intermediate potentials rely on approximate
$x_0$ reconstructions and off-the-shelf reward models, which can be noisy and introduce substantial variance.

A key next step is broader validation on established controllable-generation and safety benchmarks,
including bias/fairness suites (BOLD, HolisticBias) \citep{Dhamala_2021, smith2022imsorryhearthat},
truthfulness (TruthfulQA) \citep{lin2022truthfulqameasuringmodelsmimic},
standardized red-teaming (HarmBench) \citep{mazeika2024harmbenchstandardizedevaluationframework},
and instruction-following evaluations (MT-Bench, AlpacaEval) \citep{zheng2023judgingllmasajudgemtbenchchatbot, dubois2025lengthcontrolledalpacaevalsimpleway},
ideally within broader suites such as HELM \citep{liang2023holisticevaluationlanguagemodels}.
Future work can also test generalizability on larger discrete diffusion models such as Dream-7B \citep{ye2025dream7bdiffusionlarge} or LLaDA \citep{nie2025largelanguagediffusionmodels}. Another interesting avenue for future work is to apply the methods to sampling problems similar to \citet{wu2025practical} and exploring image steering \citep{singhal2025framework}.

\bibliography{iclr2026_conference}
\bibliographystyle{iclr2026_conference}

\newpage
\appendix

\section{Comparing SMC Algorithms}
\label{app:algorithms}

\begin{algorithm}[H]
    \caption{Bootstrap SMC}
    \label{alg:smc}
    \begin{algorithmic}
    \Require Base kernel $f$, potentials $\left\{ G^{\star}_{t} \right\}_{t\geq0}$, population sizes $N, M$
    \State Set $w_{T}^{(i)} \gets 1/N$, for $i=1,\dots,N$
    \State Sample $x_{T}^{(i)} \sim \mu(x_T) \, G^{\star}_{T}(x_{T})$
    \For{$t=T$ \textbf{to} $1$}
        \For{$i=1$ \textbf{to} $N$}
            \State Resample $a \sim \mathrm{Cat}(w_{t}^{(1:N)})$
            \State Sample $x_{t-1}^{(i)} \sim f(\cdot\mid x_{t}^{(a)})$
            \State Compute $\tilde w_{t-1}^{(i)} \gets G^\star_{t-1}(x_{t-1}^{(i)}, x_{t}^{(a^i)})$
        \EndFor
        \State Normalize $w_{t-1}^{(i)} \gets \tilde w_{t-1}^{(i)} / \sum_{k=1}^N \tilde w_{t-1}^{(k)}$
    \EndFor
    \State \Return $\{(x_0^{(i)}, w_0^{(i)})\}_{i=1}^N$
    \end{algorithmic}
\end{algorithm}

\begin{algorithm}[H]
    \caption{Fully-Adapted Nested SMC}
    \label{alg:fa-nsmc}
    \begin{algorithmic}
    \Require Base kernel $f$, potentials $\left\{ G^{\star}_{t} \right\}_{t\geq0}$, population sizes $N, M$
    \State Set $w_{T}^{(i)} \gets 1/N$, for $i=1,\dots,N$
    \State Sample $x_{T}^{(i)} \sim \mu(x_T) \, G^{\star}_{T}(x_{T})$
    \For{$t=T$ \textbf{to} $1$}
        \For{$i=1$ \textbf{to} $N$}
            \State $x_{t-1}^{(i,\cdot)}, v^{(i,\cdot)}, \hat\nu^{(i)} \gets$ \Call{NestedProposal}{$x_t^{(i)}, f, G^{\star}_{t-1}$}
        \EndFor
        \State Compute $\Omega^{(i)} \propto w_t^{(i)} \cdot \hat\nu^{(i)}$, for $i=1,\dots,N$
        \For{$i=1$ \textbf{to} $N$}
            \State Resample $a \sim \mathrm{Cat}(\Omega^{1:N})$
            \State Sample $b \sim \mathrm{Cat}\left(v^{(a,\cdot)} / \sum_k v^{(a,k)}\right)$
            \State Set $x_{t-1}^{(i)} \gets x_{t-1}^{(a,b)}$, $w_{t-1}^{(i)} \gets 1/N$
        \EndFor
    \EndFor
    \State \Return $\{(x_0^{(i)}, w_0^{(i)})\}_{i=1}^N$
    \end{algorithmic}
\end{algorithm}

\section{Derivation of Optimal Potentials}
\label{app:potential}

We now examine the ratio defining the optimal proposal $q^\star_{t-1}(x_{t-1} \mid x_t)$, which targets the intermediate distribution $\gamma_{t-1}(x_{t-1:T})$:
\begin{equation}
    q^\star_{t-1}(x_{t-1} \mid x_t) 
    = \frac{\gamma_{t-1}(x_{t-1:T})}{\gamma_t(x_{t:T})} 
    = \frac{p(x_{t-1:T}) \, \psi_{t-1}^\star(x_{t-1})}{p(x_{t:T}) \, \psi_t^\star(x_t)}.
\end{equation}
Using the Markov factorization of the prior path measure $p(x_{t-1:T}) = f(x_{t-1} \mid x_t) \, p(x_{t:T})$, this expression simplifies to:
\begin{equation}
    \frac{\gamma_{t-1}(x_{t-1:T})}{\gamma_t(x_{t:T})} = f(x_{t-1} \mid x_t) \frac{\psi_{t-1}^\star(x_{t-1})}{\psi_t^\star(x_t)} = f(x_{t-1} \mid x_t) \, G^\star_{t-1}(x_{t-1}, x_t),
\end{equation}
where we identify the ideal Feynman--Kac potential $G^\star_{t-1}$ as the ratio of expected future rewards:
\begin{equation}
    G^\star_{t-1}(x_{t-1}, x_t)
    \coloneq
    \frac{\mathbb{E}_{p}\!\left[\exp\!\big(\lambda \, r(x_0)\big) \mid x_{t-1}\right]}
         {\mathbb{E}_{p}\!\left[\exp\!\big(\lambda \, r(x_0)\big) \mid x_t\right]}.
\end{equation}

\section{Telescoping Property}
\label{app:telescoping}

Recall the optimal twisting functions $\psi_t^\star(x_t) = \mathbb{E}_{p} \big[\exp(\lambda r(x_0)) \mid x_t\big]$ and the induced potentials $G^\star_{t-1}(x_{t-1}, x_t) = \psi_{t-1}^\star(x_{t-1}) / \psi_t^\star(x_t)$. Taking the product of these potentials over the full reverse-time trajectory from $t=T$ down to $1$ yields a telescoping ratio:
\begin{align}
    \prod_{t=1}^{T} G^\star_{t-1}(x_{t-1}, x_t)
    &= \prod_{t=1}^{T} \frac{\psi_{t-1}^\star(x_{t-1})}{\psi_t^\star(x_t)} \nonumber\\
    &= \frac{\psi_0^\star(x_0)}{\psi_T^\star(x_T)}
    = \frac{\exp(\lambda \, r(x_0))}{\mathbb{E}_{p} \left[\exp(\lambda \, r(x_0)) \mid x_T\right]}.
\end{align}
The numerator is exactly the terminal tilt required by the target distribution $p_\lambda(x_0)$. The denominator depends only on the initial noise state $x_T$ and serves as the global normalizing constant. Consequently, weighting the prior path measure by this product recovers the correct target:
\begin{equation}
    p(x_{0:T}) \prod_{t=1}^{T} G^\star_{t-1}(x_{t-1}, x_t) \propto p(x_{0:T}) \exp(\lambda r(x_0)) = \gamma_0(x_{0:T}).
\end{equation}
This demonstrates that the cumulative product of the optimal potentials correctly recovers the reward-tilted posterior distribution.

\section{Intermediate Reward Plots}
\label{inter-plots}

\begin{figure}[htbp]
    \centering
    \includegraphics[width=0.95\linewidth]{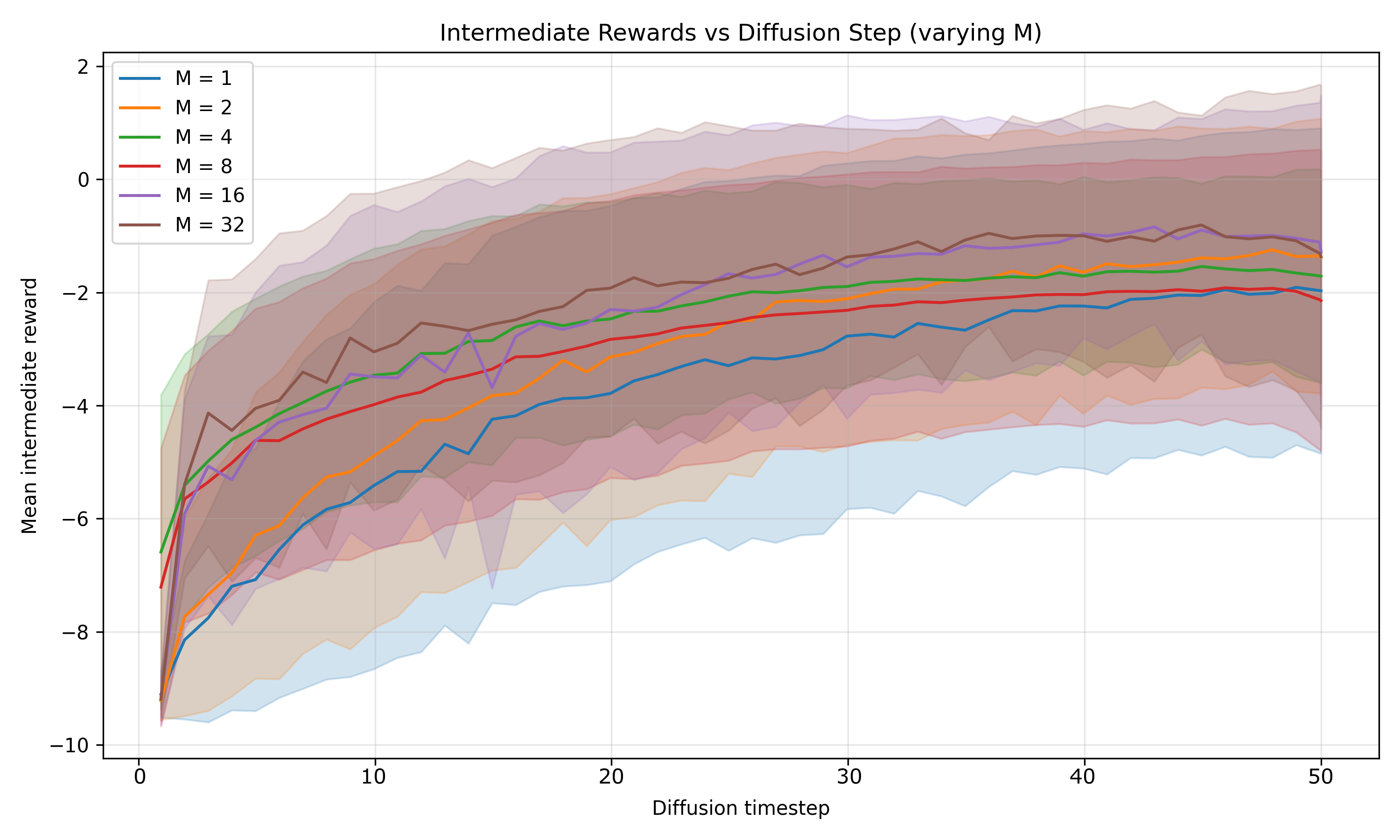}
    \caption{
    Intermediate rewards under toxicity steering increase over time, with higher population size $M$, yielding better rewards but diminishing returns and a largely stable ranking across time steps.}
    \label{fig:intermediate_rewards_by_M}
\end{figure}

\FloatBarrier

\section{Diversity}

\begin{table}[htbp]
    \centering
    \caption{Output diversity on the toxicity task (Distinct-$n$; higher is more diverse) with $N=8$, $K=4$, $\lambda=10$.}
    \label{tab:diversity}
    \begin{tabular}{lcc}
        \toprule
        Method & Distinct-1 $\uparrow$ & Distinct-2 $\uparrow$ \\
        \midrule
        BoN & 0.29 & 0.74 \\
        SMC & 0.26 & 0.71 \\ 
        NSMC & 0.26 & 0.70 \\
        FA-NSMC & 0.26 & 0.71 \\
        \bottomrule
    \end{tabular}
\end{table}

\paragraph{Output Diversity.}
We measure the diversity of the generations using Distinct-$n$ \citep{tevet-berant-2021-evaluating}. Table~\ref{tab:diversity} shows broadly comparable diversity across particle-based methods, with BoN slightly higher in this setting.

\paragraph{Diversity (Table~\ref{tab:diversity}).}
Distinct-$n$ is broadly similar across particle-based methods, with BoN slightly higher. This is consistent with resampling-induced duplication in SMC-style samplers, which can modestly reduce diversity \citep{naesseth2019smc}. However, the reverse diffusion transitions still inject substantial randomness, so we do not observe strong mode collapse in this setting despite large gains in reward attainment.

\section{Biased Nested Ablation}
\label{app:wrong_nested}

Table~\ref{tab:wrong_nested} reports the average toxicity accuracy (and external toxicity accuracy) over 10 runs for the ``biased nested'' variant discussed in the main text. Configurations vary the number of outer particles, inner proposals, and number of $x_0$ samples used for intermediate potentials; all runs use $\lambda=10$.

\begin{table}[htbp]
    \centering
    \caption{Toxicity rates for biased NSMC implementation according to \citet{uehara2025inferencetime} averaged over 10 repetitions ($\lambda=10$).}
    \label{tab:wrong_nested}
    \begin{tabular}{ccccc}
        \toprule
        $x_0$ samples & Particles & Inner particles & Toxic acc & Ext toxic acc \\
        \midrule
        4 & 4 & 4 & 0.13 & 0.13 \\
        4 & 4 & 8 & 0.17 & 0.17 \\
        4 & 8 & 4 & 0.03 & 0.01 \\
        4 & 8 & 8 & 0.20 & 0.16 \\
        16 & 4 & 4 & 0.16 & 0.13 \\
        16 & 4 & 8 & 0.26 & 0.20 \\
        16 & 8 & 4 & 0.03 & 0.01 \\
        16 & 8 & 8 & 0.30 & 0.20 \\
        \bottomrule
    \end{tabular}
\end{table}

\FloatBarrier

\end{document}